\documentclass[11pt,a4paper]{article}
\usepackage[hyperref]{emnlp2020}
\usepackage{times}
\usepackage{latexsym}

\usepackage{microtype}

\aclfinalcopy 
\def\aclpaperid{2047} 

\usepackage{times}
\usepackage{soul}
\usepackage{url}
\usepackage{hyperref}
\usepackage[utf8]{inputenc}
\usepackage{caption}
\usepackage{graphicx}
\usepackage{amsmath}
\usepackage{amsthm}
\usepackage{booktabs}
\usepackage{algorithm}
\usepackage{algorithmic}
\usepackage[utf8]{inputenc} 
\usepackage[T1]{fontenc}    
\usepackage{hyperref}       
\usepackage{url}            
\usepackage{booktabs}       
\usepackage{amsfonts}       
\usepackage{nicefrac}       
\usepackage{microtype}      
\usepackage{graphicx}
\usepackage{subfig}
\usepackage{booktabs} 
\usepackage{ amssymb }
\usepackage{amsthm}
\usepackage{amsmath}
\usepackage{arydshln}
\usepackage{xcolor}
\usepackage[utf8]{inputenc}
\usepackage{url}
\usepackage[demo]{rotating}
\usepackage{amssymb}
\usepackage{bbm}
\usepackage{rotating}
\usepackage{esvect}
\usepackage{algorithm}
\usepackage{algorithmic}

\newtheorem{defn}{Definition}
\newenvironment{definition}{\begin{defn}\em}{\end{defn}}

\newcommand{\vctr}[1]{\ensuremath{\pmb{#1}}}
\newcommand{\mtrx}[1]{\ensuremath{\pmb{#1}}}
\newcommand{\graph}[1]{\ensuremath{\mathcal{#1}}}
\newcommand{\vertices}[1]{\ensuremath{\mathcal{#1}}}

\newcommand{\vertex}[1]{\ensuremath{\mathsf{#1}}}

\newcommand{\relations}[1]{\ensuremath{\mathcal{#1}}}

\newcommand{\relation}[1]{\ensuremath{\mathsf{#1}}}

\newcommand{\function}[1]{\ensuremath{\mathtt{#1}}}

\newcommand{\Th}[1]{\ensuremath{#1^{\text{th}}}}

\newcommand{\sumElemProd}[1]{\ensuremath{\langle#1\rangle}}

\title{Out-of-Sample Representation Learning for Knowledge Graphs\thanks{~Accepted at the findings of the 2020 Conference on Empirical Methods in Natural Language Processing (EMNLP).}}

\author{Marjan Albooyeh, Rishab Goel, and Seyed Mehran Kazemi \\
    Borealis AI, Montreal, Canada \\
    \{marjan.albooyeh, rishab.goal, mehran.kazemi\}@borealisai.com}

\date{}

\begin{document}
\maketitle
\begin{abstract}
Many important problems can be formulated as reasoning in knowledge graphs. Representation learning has proved extremely effective for {\em transductive reasoning}, in which one needs to make new predictions for already observed entities. This is true for both attributed graphs (where each entity has an initial feature vector) and non-attributed graphs (where the only initial information derives from known relations with other entities). For {\em out-of-sample reasoning}, where one needs to make predictions for entities that were unseen at training time, much prior work considers attributed graph. However, this problem is surprisingly under-explored for non-attributed graphs. In this paper, we study the \emph{out-of-sample representation learning} problem for non-attributed knowledge graphs, create benchmark datasets for this task, develop several models and baselines, and provide empirical analyses and comparisons of the proposed models and baselines.
\end{abstract}

\section{Introduction}
\begin{figure}[t]
   \centering
   \resizebox{\columnwidth}{!}{%
   \includegraphics[width=\columnwidth]{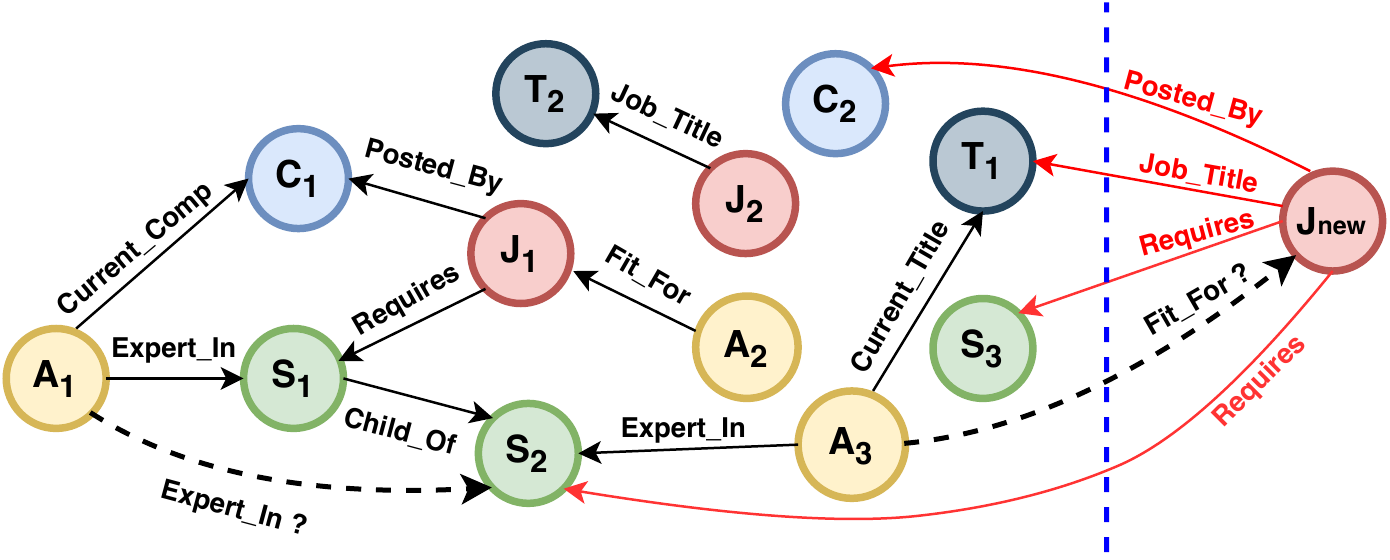}
   }
   \caption{%
   \label{fig:intro} %
   An example of a multi-relational graph for a job recommendation system is presented on the left side of the dashed blue line where the vertices $\vertex{A}_i$, $\vertex{C}_i$, $\vertex{S}_i$, $\vertex{J}_i$ and $\vertex{T}_i$ represent applicants, companies, skills, job postings, and titles respectively. Predicting whether $\vertex{A}_{1}$ is expert in $\vertex{S}_2$ is an example of transductive reasoning. $\vertex{J}_{new}$ represents an out-of-sample entity that has not been observed during training. Predicting whether $\vertex{A}_3$ is a good fit for $\vertex{J}_{new}$ based on the relations of $\vertex{J}_{new}$ observed during test time (red arrows) is an example of out-of-sample reasoning.}
\end{figure}

Multi-relational graphs are a prevalent form of graphs where each edge has a label and a direction associated with it. Many prediction problems can be formulated as reasoning within a multi-relational graph. For example, Figure~\ref{fig:intro} depicts a job recommendation system that has been formulated in these terms. 
A notable example of multi-relational graphs is knowledge graphs (KGs) with several applications in natural language processing and information retrieval including search, question answering and commonsense reasoning.
Much prior work has considered transductive KG reasoning in which predictions are made at test time for only those entities that were observed during training. These are known as {\em in-sample} entities. In Figure~\ref{fig:intro}, predicting if $\vertex{A}_1$ is expert in $\vertex{S}_2$ is an example of transductive reasoning.

Conversely, we consider \emph{out-of-sample} KG reasoning. We make predictions for previously unseen or \emph{out-of-sample} entities based on their relations with the in-sample entities.  This is more challenging than transductive reasoning as it requires generalizing to unseen entities. In Figure~\ref{fig:intro},  predicting whether $\vertex{A}_3$ is a good fit for the previously unseen job posting $\vertex{J}_{new}$ given $\vertex{J}_{new}$'s relations with in-sample entities (observed at test time) is an example of out-of-sample reasoning.  

Representation learning has proved effective for reasoning in KGs \cite{nickel2016review,hamilton2017representation,kazemi2020representation}. It has been extensively studied for transductive reasoning in {\em attributed} graphs (where each entity has an initial feature vector) and {\em non-attributed} KGs (where the only initial information derives from known relations with other entities) as well as {\em simple graphs} (in which there is only a single relation). One prominent family of work is based on extensions of the convolution operator to non-Euclidean domains \cite{kipf2017semi,defferrard2016convolutional,hammond2011wavelets,schlichtkrull2018modeling}. A second family models relations as translations (or rotations) from subject to object entities \cite{bordes2013translating,ji2015knowledge,StransE,sun2019rotate}. A third approach represents the facts in a KG as a 3rd order tensor and factorizes this tensor to produce entity and relation embeddings \cite{yang2015embedding,trouillon2016complex,kazemi2018simple,zhang2019quaternion}.

Out-of-sample representation learning has also been extensively studied for attributed KGs \cite{xie2016representation,zhao2017zero} and attributed simple graphs \cite{yang2016revisiting,hamilton2017inductive,velivckovic2018graph,chen2018fastgcn}. However, for non-attributed KGs, it remains under-explored. The main challenge of out-of-sample representation learning for non-attributed KGs is that an entity representation must be learned using only the relations the entity participates in.
\citeauthor{ma2018depthlgp} (\citeyear{ma2018depthlgp}) develop such a model for non-attributed simple graphs but extending their work to KGs is not straightforward.
Out-of-sample representation learning in non-attributed graphs is an important problem for high-throughput production systems, as it is not tractable to adapt the transductive approaches and use additional rounds of gradient descent to incorporate new entities at test time.

The contributions of this work are as follows: 1) we formally define out-of-sample representation learning for KGs, 2) we create benchmark datasets for this problem, 3) we propose several baselines, 4) we extend current transductive KG representation learning approaches by developing new training algorithms that can support the incorporation of out-of-sample entities at test time via aggregation functions to compute representations, and 5) we provide a thorough experimental comparison of the baselines and the proposed approaches.

\section{Background and Notation} \label{sec:background}
Lower-case letters denote scalars, bold lower-case letters denote vectors, and bold upper-case letters denote matrices. For a vector $\vctr{z}\in\mathbb{R}^d$, we represent by $\vctr{z}[i]$ ($n \leq d$) the $\Th{i}$ element of $\vctr{z}$ and by $||\vctr{z}||$ the Euclidean norm of $\vctr{z}$.
For $\vctr{z_1},\vctr{z_2}\in\mathbb{R}^{d}$, we let $\vctr{z}_1 \odot \vctr{z}_2\in\mathbb{R}^{d}$ represent the element-wise (Hadamard) product of the two vectors. For $\vctr{z}_1, \dots, \vctr{z}_k \in \mathbb{R}^d$, we let $\sumElemProd{\vctr{z}_1, \dots, \vctr{z}_k}=\sum_{i=1}^d (\vctr{z}_1[i] * \dots * \vctr{z}_k[i])$ represent the sum of the element-wise product of the elements of the $k$ vectors.

Let \vertices{V} and \relations{R} represent a set of entities and relations respectively. 
We represent a \textbf{triple} as  $(\vertex{v}, \relation{r}, \vertex{u})$, where $\vertex{v}\in \vertices{V}$ is the \emph{head} (or subject), $\relation{r}\in\relations{R}$ is the \emph{relation}, and $\vertex{u}\in \vertices{V}$ is the \emph{tail} (or object) of the triple. Let $\zeta$ represent the set of all triples on entities $\vertices{V}$ and relations $\relations{R}$ that are facts (e.g., $(\vertex{Montreal}, \relation{LocatedIn}, \vertex{Canada})$).
A (non-attributed) \textbf{knowledge graph (KG)}  $\graph{G} \subset \zeta$ is a subset of $\zeta$. Hereafter, whenever we refer to a KG, we assume a non-attributed KG.

\textbf{Transductive KG Reasoning:} In transductive KG reasoning, a model is learned for a KG $\graph{G}$ with entities $\vertices{V}$ and relations $\relations{R}$ such that the model can make predictions about any triple $(\vertex{v}, \relation{r}, \vertex{u})$ where $\vertex{v}, \vertex{u} \in \vertices{V}$ are both in-sample entities and $\relation{r}\in\relations{R}$.

KG embedding models map entities and relations to hidden representations known as \emph{embeddings} and define a function $\phi$ from the embeddings of the entities and the relation in a triple to a score corresponding to the degree of belief the model has for the relation holding between the entities. Typically, the embeddings can be formulated as two matrices $\mtrx{Z}_{ent}\in\mathbb{R}^{|\vertices{V}|\times d_{ent}}$ and $\mtrx{Z}_{rel}\in\mathbb{R}^{|\relations{R}|\times d_{rel}}$ where each row of $\mtrx{Z}_{ent}$ corresponds to the embedding for an entity, each row of $\mtrx{Z}_{rel}$ corresponds to the embedding for a relation, and $d_{ent}$ and $d_{rel}$ represent entity and relation embedding sizes. One can look up the embedding for a particular entity $\vertex{v}$ by multiplying the transpose of $\mtrx{Z}_{ent}$ to the one-hot encoding of $\vertex{v}$ and for a particular relation $\vertex{r}$ by multiplying the transpose of $\mtrx{Z}_{rel}$ to the one-hot encoding of $\relation{r}$. A large number of approaches define $\mtrx{Z}_{ent}$ and $\mtrx{Z}_{rel}$ as matrices with directly learnable parameters. Other approaches define encoders that produce these two matrices typically through several rounds of message passing among entities. 

Algorithm~\ref{algo:training} outlines one epoch of training for learning the embeddings as well as the parameters of the $\phi$ function. The training is performed using stochastic gradient descent with mini-batches. For each batch (line \ref{in-algo:batch}), the $\function{nextBatch}$ function extracts a set of positive triples from the KG and creates $n$ negative triples per positive triple by corrupting the positive triple according to the procedure introduced in \cite{bordes2013translating}. $n$ is known as the \emph{negative ratio}. For each triple $(\vertex{v}, \relation{r}, \vertex{u})$ in the batch, the embeddings for $\vertex{v}, \relation{r}$ and $\vertex{u}$ are looked up and the score for the triple is computed according to $\phi$. Then the embeddings and the parameters of $\phi$ are updated based on the predicted scores, the labels of the triples, and a loss function $\mathcal{L}$.

\begin{algorithm}[t]
\caption{Transductive Training (one epoch)}
\label{algo:training} \textbf{Inputs} $n:$ negative ratio, $\mathcal{L}:$ loss function

\begin{algorithmic}[1]
\FOR{$batch=1$ {\bfseries to} $numBatches$}
    \STATE $triples, labels$ $\leftarrow$
    \function{nextBatch}\,($batch$, $n$) \label{in-algo:batch}
    \STATE $scores \leftarrow []$
    \FOR{$(\vertex{v}, \relation{r}, \vertex{u})$ {\bfseries in} $triples$}
        \STATE $\vctr{z}_\vertex{v} \leftarrow \function{lookup}(\vertex{v}, \mtrx{Z}_{ent})$
        \STATE $\vctr{z}_\relation{r} \leftarrow \function{lookup}(\relation{r}, \mtrx{Z}_{rel})$
        \STATE $\vctr{z}_\vertex{u} \leftarrow \function{lookup}(\vertex{u}, \mtrx{Z}_\vertex{ent})$
        \STATE $scores$.\function{append}($\phi(\vctr{z}_\vertex{v}, \vctr{z}_\relation{r}, \vctr{z}_\vertex{u})$)
    \ENDFOR
    \STATE \function{updateParams}($\mathcal{L}$, $scores$, $labels$)
\ENDFOR
\end{algorithmic}
\end{algorithm}

Different models have been proposed in the literature by mainly changing the score function. Note that some models may break the vector embeddings into multiple pieces and reshape each piece before using it in the score function. In this paper, we focus primarily on DistMult, a simple yet effective model for transductive KG embedding. However, many of the ideas we develop in this paper are general and can be applied to other models as well.

\textbf{DistMult \cite{yang2015embedding}:} In DistMult, $\mtrx{Z}_{ent}\in\mathbb{R}^{|\vertices{V}|\times d}$ and $\mtrx{Z}_{rel}\in\mathbb{R}^{|\relations{R}|\times d}$. For a triple $(\vertex{v}, \relation{r}, \vertex{u})$, let $\vctr{z}_\vertex{v}, \vctr{z}_\relation{r}, \vctr{z}_\vertex{u} \in \mathbb{R}^d$ represent the embeddings for $\vertex{v}$, $\relation{r}$ and $\vertex{u}$ respectively where each embedding is obtained by looking up the $\mtrx{Z}_{ent}$ and $\mtrx{Z}_{rel}$ matrices. DistMult defines the score for the triple as $\phi(\vctr{z}_\vertex{v}, \vctr{z}_\relation{r}, \vctr{z}_\vertex{u})=\sumElemProd{\vctr{z}_\vertex{v}, \vctr{z}_\relation{r}, \vctr{z}_\vertex{u}}$, i.e. the sum of the element-wise product of the head, relation, and tail embeddings.

\textbf{Loss function:} We use the L2 regularized negative log-likelihood which has proved effective in several works \cite{trouillon2016complex,kazemi2018simple}. The loss $\mathcal{L}(\Theta)$ for a single batch of labeled triples is defined as follows:
\begin{equation}
     \sum_{((\vertex{v}, \relation{r}, \vertex{u}), l) \in batch} \function{softplus}(-l~\cdot~\phi(\vertex{v}, \relation{r}, \vertex{u})) + \lambda ||\Theta||^2_2
\end{equation}
where $\Theta$ represents the parameters of the model, $\function{softplus}(x)=\function{log}(1+\function{exp}(x))$, $l\in\{-1,1\}$ represents the label of the triple in the batch, and $\lambda$ represents the L2 regularization hyperparameter.

\section{Out-of-Sample KG Reasoning} \label{sec:out-of-sample}
We define out-of-sample reasoning for KGs as:

\begin{definition} \label{dfnt:out-of-sample}
Out-of-sample reasoning for KGs is the problem of training a model on a KG $\graph{G}$ with entities $\vertices{V}$ and relations $\relations{R}$ such that at the test time, the model can be used for making predictions about any out-of-sample entity $\vertex{v}\not\in\vertices{V}$ given $\graph{G}_\vertex{v}=\{(\vertex{v}, \relation{r}, \vertex{u}): \vertex{u}\in\vertices{V}, \relation{r}\in\relations{R}\}\cup\{(\vertex{u}, \relation{r}, \vertex{v}): \vertex{u}\in\vertices{V}, \relation{r}\in\relations{R}\}$ corresponding to the relations between $\vertex{v}$ and in-sample entities.
\end{definition}

According to the definition, $\graph{G}_\vertex{v}$ is observed only at the test time and so during training, the model does not observe any triples involving $\vertex{v}$.
To develop a representation learning model for out-of-sample reasoning in KGs, one needs to learn i) embeddings for the in-sample entities in $\vertices{V}$ and the relations in $\relations{R}$, ii) a function $\phi$ from triples to scores, and iii) a function from $\graph{G}_\vertex{v}$ and the in-sample entity and relation embeddings to an embedding for $\vertex{v}$ that can be used to make further predictions about $\vertex{v}$.

One possible way of extending transductive models such as DistMult to the out-of-sample domain is by following the standard training procedure outlined in Algorithm~\ref{algo:training} and then defining an aggregation function with no learnable parameters which, at inference time, provides an embedding for an out-of-sample entity $\vertex{v}$ based on the embeddings of the entities and relations in $\graph{G}_\vertex{v}$. A simple aggregation function, for instance, can be the average of the embeddings for entities $\{\vertex{u}: \exists \relation{r}~s.t.~ (\vertex{v}, \relation{r}, \vertex{u})\in\graph{G}_\vertex{v}~or~(\vertex{u}, \relation{r}, \vertex{v})\in\graph{G}_\vertex{v}\}$ (i.e. all entities that have a relation with $\vertex{v}$). Such a procedure, however, introduces an inconsistency between training and testing as the training is done irrespective of the aggregation function and with the objective of performing well on a transductive task whereas the model is tested on an out-of-sample task. 

\subsection{Proposed Training Procedure} \label{sec:new-train-proc}
To make the training procedure resemble what is expected of the model at the test time and make it aware of the aggregation function being used, we propose a new training algorithm that guides the learning procedure towards learning entity and relation embeddings that better match the aggregation function.
A general training procedure for out-of-sample representation learning is proposed in Algorithm~\ref{algo:ind-training}. For each triple $(\vertex{v}, \relation{r}, \vertex{u})$ in the batch, first we lookup the embedding for $\relation{r}$. Then with probability $\frac{\psi}{2}$, where $0\leq\psi\leq 1$ is a hyperparameter, we consider $\vertex{v}$ to be out-of-sample and $\vertex{u}$ to be in-sample. In this case, for $\vertex{v}$ we use an aggregate function that computes the embedding for $\vertex{v}$ based on the triples involving $\vertex{v}$ except for $(\vertex{v}, \relation{r}, \vertex{u})$, and for $\vertex{u}$ we simply lookup its embedding. Also with probability $\frac{\psi}{2}$, we consider $\vertex{u}$ to be out-of-sample and $\vertex{v}$ to be in-sample and follow a similar procedure. Finally, with probability $1-\psi$, we follow the standard training procedure by looking up the embedding for both entities. Having the embeddings for $\vertex{v}$, $\relation{r}$ and $\vertex{u}$, we use a score function (e.g., DistMult) to compute the score for this triple being true. Finally, we update the embeddings (and the parameters of the aggregate and $\phi$ functions if they have any) according to the scores, labels, and a loss function $\mathcal{L}$. Note that when $\psi=0$, Algorithm~\ref{algo:ind-training} reduces to Algorithm~\ref{algo:training}. Note that Algorithm~\ref{algo:ind-training} is generic and can be used with any KG embedding model.

\begin{algorithm}[t]
\caption{Out-of-Sample Training (one epoch)}
\label{algo:ind-training}
\textbf{Inputs~} $n:$ negative ratio, $\mathcal{L}:$ loss function, $\psi:$ see Section~\ref{sec:new-train-proc}

\begin{algorithmic}[1]

\FOR{$batch=1$ {\bfseries to} $numBatches$}
    \STATE $triples, labels \leftarrow$ \function{nextBatch}\,($batch$, $n$)
    \STATE $scores \leftarrow []$
    \FOR{$(\vertex{v}, \relation{r}, \vertex{u})$ {\bfseries in} $triples$}
        \STATE $\vctr{z}_\relation{r} \leftarrow \function{lookup}(\relation{r}, \mtrx{Z}_\relation{r})$
        \STATE $rand \leftarrow \function{random}()$
        \IF{$rand < \frac{\psi}{2}$}
            \STATE $\vctr{z}_\vertex{v} \leftarrow \function{aggregate}(\vertex{v}, \mtrx{Z}_{rels}, \mtrx{Z}_{ent})$
            \STATE $\vctr{z}_\vertex{u} \leftarrow \function{lookup}(\vertex{u}, \mtrx{Z}_{ent})$
        \ELSIF{$\frac{\psi}{2} < rand < \psi$}
            \STATE $\vctr{z}_\vertex{v} \leftarrow \function{lookup}(\vertex{v}, \mtrx{Z}_{ent})$
            \STATE $\vctr{z}_\vertex{u} \leftarrow \function{aggregate}(\vertex{u}, \mtrx{Z}_{rel}, \mtrx{Z}_{ent})$
        \ELSE{}
            \STATE $\vctr{z}_\vertex{v} \leftarrow \function{lookup}(\vertex{v}, \mtrx{Z}_{ent})$
            \STATE $\vctr{z}_\vertex{u} \leftarrow \function{lookup}(\vertex{u}, \mtrx{Z}_{ent})$
        \ENDIF
        \STATE $scores$.\function{append}($\phi(\vctr{z}_\vertex{v}, \vctr{z}_\relation{r}, \vctr{z}_\vertex{u})$)
    \ENDFOR
    \STATE \function{updateParams}($\mathcal{L}$, $scores$, $labels$)
\ENDFOR
\end{algorithmic}
\end{algorithm}

By using Algorithm~\ref{algo:ind-training}, one can develop different models for out-of-sample representation learning by choosing different $\phi$ and $\function{aggregate}$ functions.  
We propose two aggregate functions that extend DistMult to out-of-sample domains.

\begin{table*}[t]
\begin{center}
\resizebox{\textwidth}{!}{%
\begin{tabular}{c||c|c|c|c|c|c}
 & In-sample entities &  &  & Train & Validation & Test \\
Dataset & ($|\vertices{V}|$) & Out-of-sample entities & $|\relations{R}|$ & triples & queries & queries \\ \hline
oWN18RR & 32270 & validation: 2848, test: 2848 & 11 & 60608 & 12760 & 12440 \\
oFB15k-237 & 11579 & validation: 1395, test: 1396 & 234 & 193490 & 44601 & 54082
\end{tabular}
}
\caption{\label{tab:dataset-stats}
Statistics on oWN18RR and oFB15k-237.}
\end{center}
\end{table*}

\subsection{Proposed Models}
\paragraph{oDistMult-ERAvg:} Let $\vertex{v}$ be an entity for which we need to compute an embedding using aggregation and $\graph{G}_\vertex{v}$ be the triples involving $\vertex{v}$. According to the score function of DistMult, for each triple $(\vertex{v}, \relation{r}, \vertex{u}) \in \graph{G}_\vertex{v}$ (and similarly for each triple $(\vertex{u}, \relation{r}, \vertex{v}) \in \graph{G}_\vertex{v}$), we want $\sumElemProd{\vctr{z}_\vertex{v}, \vctr{z}_\relation{r}, \vctr{z}_\vertex{u}}$ to be high where $\vctr{z}_\vertex{v}$, $\vctr{z}_\relation{r}$ and $\vctr{z}_\vertex{u}$ represent the embedding of $\vertex{v}$, $\relation{r}$ and $\vertex{u}$ respectively. The score can be written as $\sumElemProd{\vctr{z}_\vertex{v}, \vctr{z}_\relation{r}, \vctr{z}_\vertex{u}}=\vctr{z}_\vertex{v} \cdot (\vctr{z}_\relation{r} \odot \vctr{z}_\vertex{u})$ where $\cdot$ represents dot product. Since $\vctr{z}_\vertex{v} \cdot (\vctr{z}_\relation{r} \odot \vctr{z}_\vertex{u})=||\vctr{z}_\vertex{v}||~||\vctr{z}_\relation{r} \odot \vctr{z}_\vertex{u}||~\function{cos}(\vctr{z}_\vertex{v}, \vctr{z}_\relation{r} \odot \vctr{z}_\vertex{u})$, one possible choice to ensure a high value for $\sumElemProd{\vctr{z}_\vertex{v}, \vctr{z}_\relation{r}, \vctr{z}_\vertex{u}}$ is by choosing $\vctr{z}_\vertex{v}$ to be the vector $\vctr{z}_\relation{r} \odot \vctr{z}_\vertex{u}$ so that the angle $\theta$ between the two vectors becomes $0$ (and consequently, $\function{cos}(\theta)=1$). Since there may be multiple triples in $\graph{G}_\vertex{v}$, we average these vectors and define $\vctr{z}_\vertex{v} = \function{aggregate}(\vertex{v})$ as follows:
\begin{equation}
\label{eq:er-avg}
\vctr{z}_\vertex{v}=\frac{1}{|\graph{G}_\vertex{v}|}(\sum_{(\vertex{v}, \relation{r}, \vertex{u})\in\graph{G}_\vertex{v}} \vctr{z}_\relation{r} \odot \vctr{z}_\vertex{u} + \sum_{(\vertex{u}, \relation{r}, \vertex{v})\in\graph{G}_\vertex{v}} \vctr{z}_\relation{r} \odot \vctr{z}_\vertex{u})
\end{equation}
where $|\graph{G}_\vertex{v}|$ represents the number of triples in $\graph{G}_\vertex{v}$.

\paragraph{oDistMult-LS:}
An alternative to the averaging strategy in Equation~\eqref{eq:er-avg} is to find $\vctr{z}_\vertex{v}$ as the solution to a least squares problem to ensure the score for the triples in $\graph{G}_\vertex{v}$ are maximized. One way to achieve this goal is by solving a (potentially under-determined) system of linear equations where there exists one equation of the form $\frac{\vctr{z}_\vertex{v} \cdot (\vctr{z}_\relation{r} \odot \vctr{z}_\vertex{u})}{||\vctr{z}_\vertex{v}||~||\vctr{z}_\relation{r} \odot \vctr{z}_\vertex{u}||}=1$
for each triple $(\vertex{v}, \relation{r}, \vertex{u})\in \graph{G}_\vertex{v}$ (and similarly for each triple $(\vertex{u}, \relation{r}, \vertex{v})\in \graph{G}_\vertex{v}$). The presence of $||\vctr{z}_\vertex{v}||$ in the denominator makes finding an analytical solution difficult. We note that $||\vctr{z}_\vertex{v}||$ only affects the magnitude of the scores and not their ranking, so instead we consider the following equation:
\begin{equation} \label{eq:ls}
\frac{\vctr{z}_\vertex{v} \cdot (\vctr{z}_\relation{r} \odot \vctr{z}_\vertex{u})}{||\vctr{z}_\relation{r} \odot \vctr{z}_\vertex{u}||}=1
\end{equation}
Considering a matrix $\mtrx{A}\in\mathbb{R}^{|\graph{G}_\vertex{v}|\times d}$ (recall that $d$ is the embedding dimension) such that $\mtrx{A}[i]=\vctr{z}_\relation{r} \odot \vctr{z}_\vertex{u}$ where $\relation{r}$ and $\vertex{u}$ are the relation and entity involved in the $i$-th triple in $\graph{G}_\vertex{v}$ and a vector $\vctr{b}\in\mathbb{R}^{|\graph{G}_\vertex{v}|}$ such that $\vctr{b}[i]=||\vctr{z}_\relation{r} \odot \vctr{z}_\vertex{u}||$, we compute $\vctr{z}_\vertex{v} = \function{aggregate}(\vertex{v})$ analytically as follows:
\begin{equation} \label{eq:least-squares}
    \vctr{z}_\vertex{v}=(\mtrx{A}^T\mtrx{A} + \lambda \mtrx{I})^{-1} \mtrx{A}^T \vctr{b}
\end{equation}
where $\mtrx{I}\in\mathbb{R}^{d\times d}$ is an identity matrix and $\lambda$ is a hyperparameter corresponding to L2 regularization which ensures the system has a unique solution.

While we proposed the aggregation functions for DistMult, note that they can be easily extended to other models such as SimplE, ComplEx, and QuatE that have 2, 4 and 8 $\sumElemProd{., ., .}$ terms respectively.

\subsection{Time Complexity}
We analyze the time complexity of the proposed algorithms for finding the embedding of an out-of-sample entity $\vertex{v}$. Let us assume that $|\graph{G}_\vertex{v}|=N$ and the embedding dimension is $d$. Finding the embedding for $\vertex{v}$ in oDistMult-ERAvg has a time complexity of $O(Nd)$ as it requires computing $N$ Hadamard products and then averaging the resulting vectors both having a time complexity of $O(Nd)$. 

For oDistMult-LS, to create the matrix $\mtrx{A}$ and vector $\vctr{b}$ one needs to compute $N$ Hadamard products and find the norm of $N$ vectors respectively. The time complexity of this step is $O(Nd)$. The size of the matrix $\mtrx{A}$ is $N\times d$ so computing $\mtrx{A}^T\mtrx{A}$ has a time complexity of $O(Nd^2)$, the matrix inversion has a time complexity of $O(d^3)$ and the product of the resulting inverted matrix into $\mtrx{A}^T$ also has a time complexity of $O(Nd^2)$. Therefore, the overall time complexity is $O(Nd^2 + d^3)$. Unless the degree size of the KG is quite large, one can expect $d$ to be larger than $N$ and so the time complexity becomes $O(d^3)$.

\section{Datasets}
We created datasets for out-of-sample representation learning over KGs using WN18RR \cite{dettmers2018convolutional} and FB15k-237 \cite{toutanova2015observed}, two standard datasets for KG completion. WN18RR is a subset of Wordnet \cite{miller1995wordnet} and FB15k-237 is a subset of Freebase \cite{bollacker2008freebase}. We call the two datasets oWN18RR and oFB15k-237 respectively, where ``o'' in the beginning of the name stands for ``out-of-sample''. The statistics for these datasets can be found in Table~\ref{tab:dataset-stats}.

We outline the steps we took for creating the datasets. 
\begin{enumerate}
    \item We merge the train, validation, and test triples from the original dataset into a single set.
    \item From the entities appearing in at least $2$ triples, we randomly select $20\%$ to be candidates for the out-of-sample entities; other entities are in-sample entities. We avoid having entities appearing in only $1$ triple as out-of-sample entities because, during test time, we select one triple as query and need other triples for learning a representation for the out-of-sample entity.
    \item \label{step:separate} Triples containing two out-of-sample entities are removed, triples with one out-of-sample entity are considered as test triples and other triples are considered as train triples.
    \item In step~\ref{step:separate}, it is possible that some entities selected to be in-sample appear in no training triples. This can happen whenever an in-sample entity only appears in triples involving an out-of-sample entity. A similar situation can occur for some relations as well (i.e. some relations only appearing in the test set). We remove such entities and relations and the triples they appear in from the dataset.
    \item After doing the above steps, if the number of triples for an out-of-sample entity is less than 2, we remove that entity from the test set.
    \item We randomly select half of the out-of-sample entities and the triples they appear in as the validation set and the other half as the test set.
\end{enumerate}

\begin{table*}[t]
\begin{center}
\resizebox{\textwidth}{!}{%
\begin{tabular}{cccccccccc}
\toprule
&  & \multicolumn{4}{c}{oWN18RR} & \multicolumn{4}{c}{oFB15k-237}                   \\
\cmidrule(lr){3-6} \cmidrule(lr){7-10}
& & \multicolumn{1}{c}{MRR} & \multicolumn{3}{c}{Hit@} & \multicolumn{1}{c}{MRR} & \multicolumn{3}{c}{Hit@} \\
\cmidrule(lr){3-3} \cmidrule(lr){4-6} \cmidrule(lr){7-7} \cmidrule(lr){8-10}
Model & Training & Filtered & 1 & 3 & 10 & Filtered & 1 & 3 & 10 \\ \hline
Popularity & Algorithm~\ref{algo:training} & 0.0094 & 0.0030 & 0.0076 & 0.0215 & 0.0320 & 0.0168 & 0.0322 & 0.0581\\
OOV & Algorithm~\ref{algo:training} & 0.0004 & 0.0000 & 0.0001 & 0.0002 & 0.0002 & 0.0000 & 0.0000 & 0.0001\\
RGCN-D & Algorithm~\ref{algo:training} & 0.0178 & 0.0072 & 0.0166 & 0.0352 & 0.1683 & 0.0974 & 0.1848 & 0.3056\\
DistMult-EAvg & Algorithm~\ref{algo:training} & 0.0446 & 0.0248 & 0.0469 & 0.0841 & 0.0813 & 0.0525 & 0.0973 & 0.1327 \\
DistMult-ERAvg & Algorithm~\ref{algo:training} & 0.3048 & 0.2468 & 0.3331 & 0.4159 & 0.2456 & 0.1615 & 0.2769 & 0.4082\\
DistMult-LS & Algorithm~\ref{algo:training} & 0.3514 & 0.2840 & 0.3911 & 0.4756 & 0.2073 & 0.1395 & 0.2264 & 0.3375 \\
DistMult-LS-U & Algorithm~\ref{algo:training} & 0.3238 & 0.2458 & 0.3693 & 0.4717 & 0.1674 & 0.1099 & 0.1858 & 0.2732\\
oDistMult-EAvg & Algorithm~\ref{algo:ind-training} & 0.2239 & 0.1315 &  0.2724 & 0.3897 & 0.1765 &  0.0724 & 0.2076 & 0.4012 \\
\hline
oDistMult-ERAvg & Algorithm~\ref{algo:ind-training} & 0.3904 & 0.3460 & 0.4125 & 0.4725 & \textbf{0.2557} & \textbf{0.1698} & \textbf{0.2885} & \textbf{0.4201}\\
oDistMult-LS & Algorithm~\ref{algo:ind-training} & \textbf{0.4093} & \textbf{0.3643} & \textbf{0.4371} & \textbf{0.4892} & 0.2126 & 0.1232 & 0.2404 & 0.3954\\
\end{tabular}
}
\caption{\label{results-table} Results on oWN18RR and oFB15k-237. Best results are in bold.}
\end{center}
\end{table*}

\section{Experiments and results}
\begin{figure*}[t]
   \centering
   \subfloat[]{%
   \includegraphics[width=0.35\textwidth]{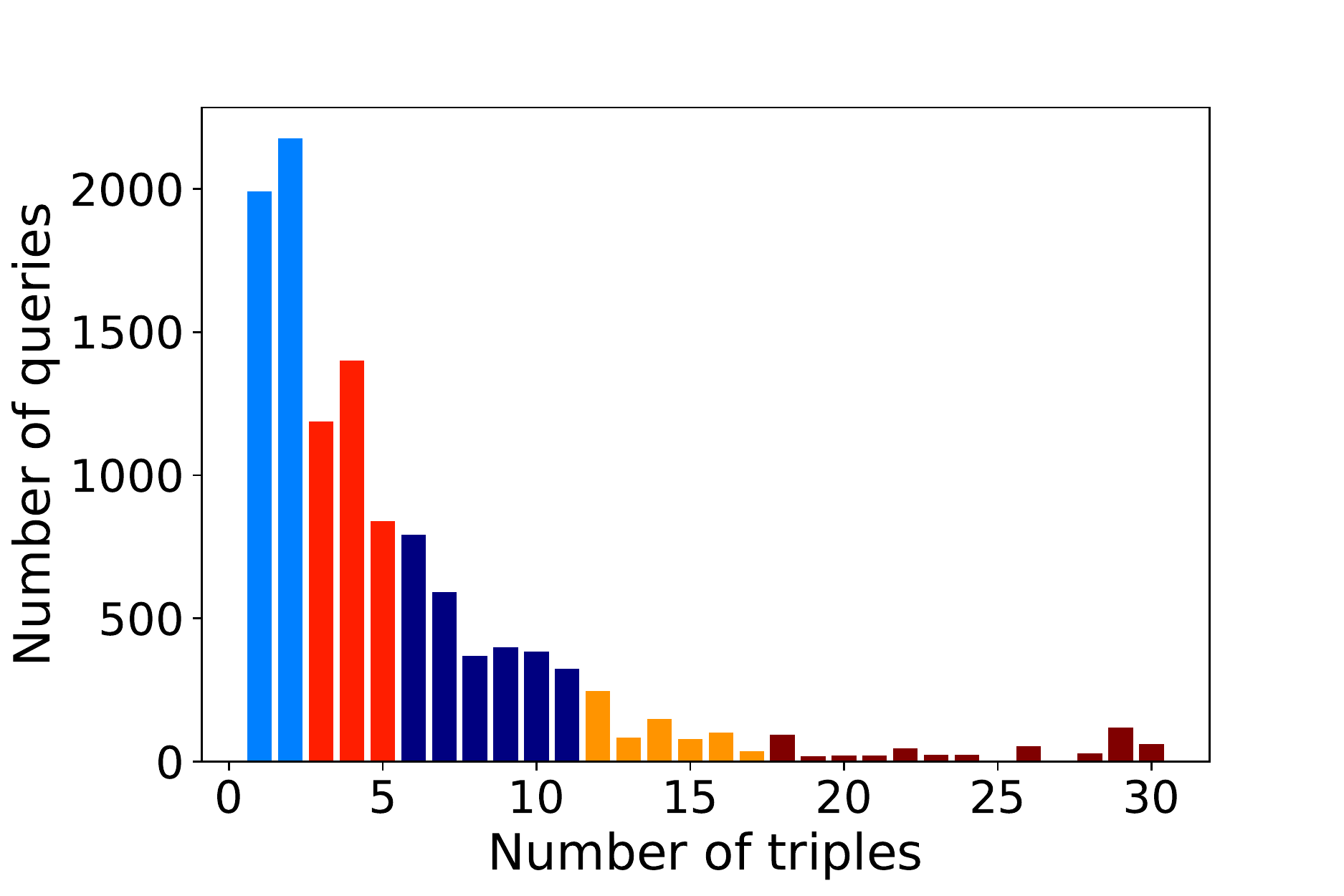}}
~~~~\hspace*{0cm}
   \subfloat[]{%
   \includegraphics[width=0.35\textwidth]{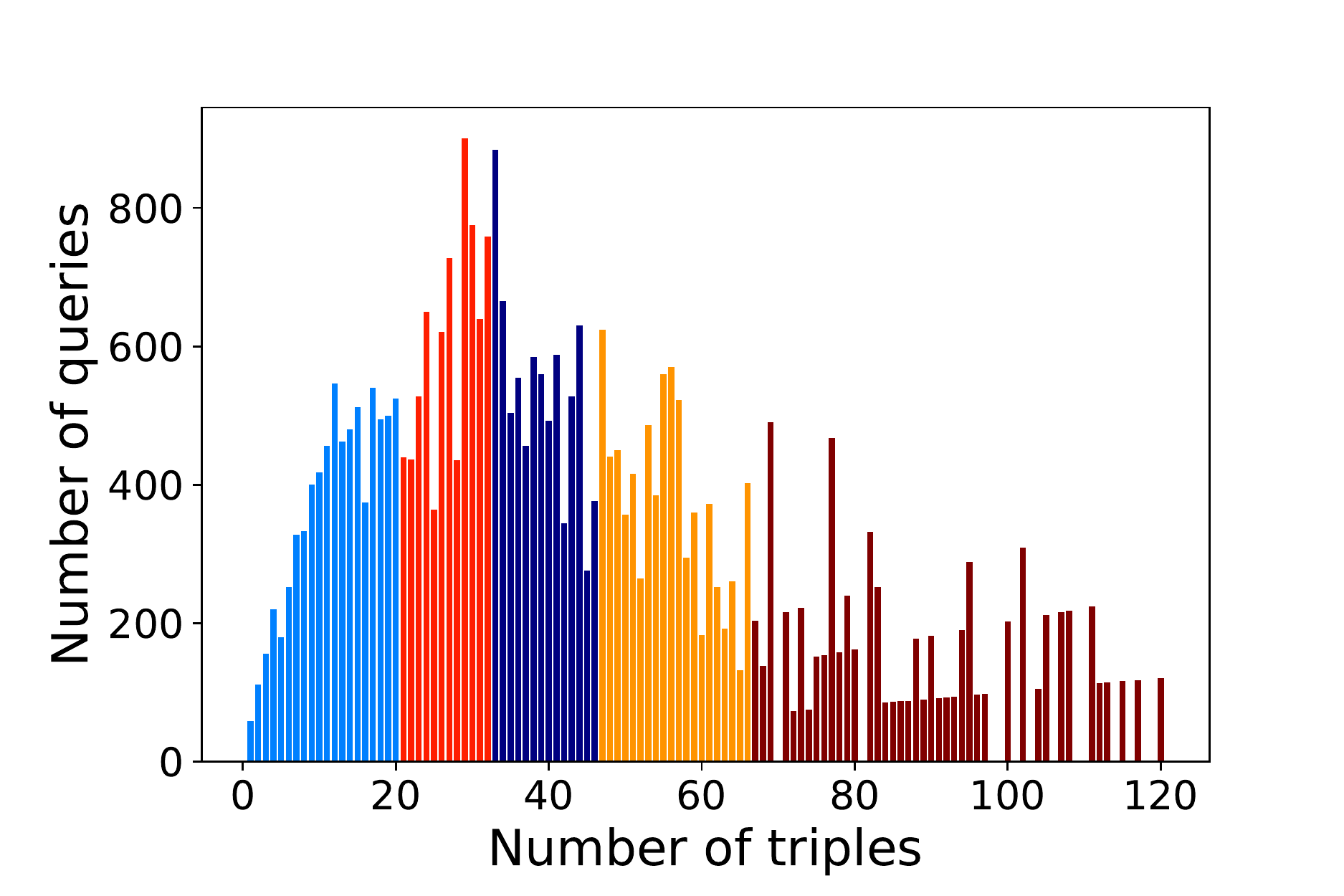}} %

   \caption{%
   \label{fig:dataset-neighbours} %
   The two figures provide statistics on the test sets of (a) oWN18RR and (b) oFB15k-237. They show the number of test queries (on the y-axis) for which the embedding of the out-of-sample entity is computed based on $k$  triples (e.g., for almost 2000 queries in oWN18RR, the embedding of the out-of-sample entity is learned based on only 1 triple).
   Since the number of samples for many of the larger values of $k$ is $0$, to make the plots visually appealing, we restricted the x-axis to $k\leq 30$ for oWN18RR and $k\leq 120$ for oFB15k-237 and did not include in the diagrams the few cases where $k$ was larger. 
   The colors show the bins used for the experiment in Figure~\ref{fig:psi-bin-results}(b, c).}
\end{figure*}

To measure the performance of different models, for any out-of-sample entity $\vertex{v}$ in the test set with triples $\graph{G}_\vertex{v}$, we create $|\graph{G}_\vertex{v}|$ queries where in the $i$-th query, we use our learned model to compute an embedding for $\vertex{v}$ given all except the $i$-th triple in $\graph{G}_\vertex{v}$ and use that embedding to make a prediction about the $i$-th triple. Figure~\ref{fig:dataset-neighbours} represents statistics on the number of triples used to compute the embedding of the out-of-sample entities in the test set for both oWN18RR and oFB15k-237. If the $i$-th triple is of the form $(\vertex{v}, \relation{r}, \vertex{u})$, then we create the query $(\vertex{v}, \relation{r}, ?)$ and find the ranking our model assigns to $\vertex{u}$ (the correct answer to the query) among entities $\vertex{u'}\in\vertices{V}$ such that $(\vertex{v}, \relation{r}, \vertex{u'})\not\in\graph{G}_\vertex{v}$ (the $(\vertex{v}, \relation{r}, \vertex{u'})\not\in\graph{G}_\vertex{v}$ constraint is known as the \emph{filtered} setting).
We follow a similar procedure for the case where the $i$-th triple is of the form $(\vertex{u}, \relation{r}, \vertex{v})$. Let $\kappa_{(\vertex{v}, \relation{r}, ?), \vertex{u}}$ represent the rank of $\vertex{u}$ for query $(\vertex{v}, \relation{r}, ?)$. 
We report filtered \emph{mean reciprocal rank (MRR)} computed as:
\begin{align}
     \frac{1}{\sum_{\vertex{v}\in Test}|\graph{G}_\vertex{v}|}\sum_{\vertex{v}\in Test}(\sum_{(\vertex{v}, \relation{r}, \vertex{u})\in\graph{G}_\vertex{v}} \frac{1}{\kappa_{(\vertex{v}, \relation{r}, ?), \vertex{u}}}+\\
     \sum_{(\vertex{u}, \relation{r}, \vertex{v})\in\graph{G}_\vertex{v}} \frac{1}{\kappa_{(?, \relation{r}, \vertex{v}), \vertex{u}}})\notag
\end{align}
and filtered Hit@k (for $k\in\{1, 3, 10\}$) defined as:
\begin{align}
    \frac{1}{\sum_{\vertex{v}\in Test}|\graph{G}_\vertex{v}|}\sum_{\vertex{v}\in Test}(\sum_{(\vertex{v}, \relation{r}, \vertex{u})\in\graph{G}_\vertex{v}} \mathbbm{1}_{\kappa_{(\vertex{v}, \relation{r}, ?), \vertex{u}}\leq k}+\\
    \sum_{(\vertex{u}, \relation{r}, \vertex{v})\in\graph{G}_\vertex{v}} \mathbbm{1}_{\kappa_{(?, \relation{r}, \vertex{v}), \vertex{u}}\leq k})\notag
\end{align}
where $\mathbbm{1}_{condition}$ is $1$ if the condition holds and $0$ otherwise.

\subsection{Baselines}
We develop several baselines for out-of-sample representation learning over KGs. 

\textbf{Popularity:} In this baseline, we rank the in-sample entities based on the number of times they appear in the triples of the training set. We break ties randomly. At the test time, we use this ranking as our answer to all queries. 

\textbf{OOV:} This baseline is inspired by the way a word embedding is computed for out-of-vocabulary (OOV) words (i.e. words unseen during training) in some works in the natural language processing literature. After training, we compute the average embedding of all in-sample entities and use it as the embedding for out-of-sample entities.

\textbf{RGCN-D:} Graph convolutional networks (GCNs) have proved effective for inductive and out-of-sample learning when initial entity features are available. When such features are not available, \citeauthor{hamilton2017inductive} (\citeyear{hamilton2017inductive}) propose to use node degrees as initial entity features. Since we work with multi-relational graphs, we initialize entity features as vectors of size $2|\relations{R}|$ where the $i$-th and $|\relations{R}|+i$-th elements (for $i < |\relations{R}|$) represent the number of incoming and outgoing edges with relation type $\relation{r}_i$ respectively. We use RGCN \cite{schlichtkrull2018modeling} as the GCN.
    
\textbf{oDistMult-EAvg:} Similar to the first baseline in \cite{ma2018depthlgp}, we create a simpler version of oDistMult-ERAvg by defining the embedding for an unseen entity $\vertex{v}$ as the average of the embeddings of the entities that are related to $\vertex{v}$. More formally, this baseline defines $\vctr{z}_\vertex{v}=\function{aggregate}(\vertex{v})=\frac{1}{|\graph{G}_\vertex{v}|}(\sum_{(\vertex{v}, \relation{r}, \vertex{u})\in\graph{G}_\vertex{v}} \vctr{z}_\vertex{u} + \sum_{(\vertex{u}, \relation{r}, \vertex{v})\in\graph{G}_\vertex{v}} \vctr{z}_\vertex{u})$.\\

\textbf{DistMult-EAvg, DistMult-ERAvg, DistMult-LS:} Corresponding to variants of oDistMult-EAvg, oDistMult-ERAvg and oDistMult-LS where instead of using Algorithm~\ref{algo:ind-training} for training, the standard training in Algorithm~\ref{algo:training} is used.

\textbf{DistMult-LS-U:} As an ablation study, we also include an unnormalized version of DistMult-LS where we change Equation~\eqref{eq:ls} to $\vctr{z}_\vertex{v} \cdot (\vctr{z}_\relation{r} \odot \vctr{z}_\vertex{u})=1$ (in other words, setting the elements of $\vctr{b}$ in Equation~\eqref{eq:least-squares} to 1).

\subsection{Implementation Details}
For RGCN-D, we used the implementation in the deep graph library (DGL).
We implemented other models and baselines in PyTorch \cite{paszke2017automatic} and used the AdaGrad optimizer \cite{duchi2011adaptive}. We selected the hyperparameters corresponding to learning rate and L2 regularization ($\lambda$) via a grid search over $\{0.1, 0.01\}$ and $\{0.1, 0.01, 0.001, 0.0001\}$ respectively validating the models every $100$ epochs and selecting the best hyperparameters and epoch based on validation filtered MRR. We set the negative ratio to $1$ and the embedding dimension to $200$. When using Algorithm~\ref{algo:ind-training} for training, we set $\psi$ to $0.5$ unless stated otherwise. The code and datasets are available at  \href{https://github.com/BorealisAI/OOS-KGE}{https://github.com/BorealisAI/OOS-KGE}.

\begin{figure*}[t]
   \centering
   \resizebox{\textwidth}{!}{%
   \subfloat[]{%
   \includegraphics[width=0.3\textwidth]{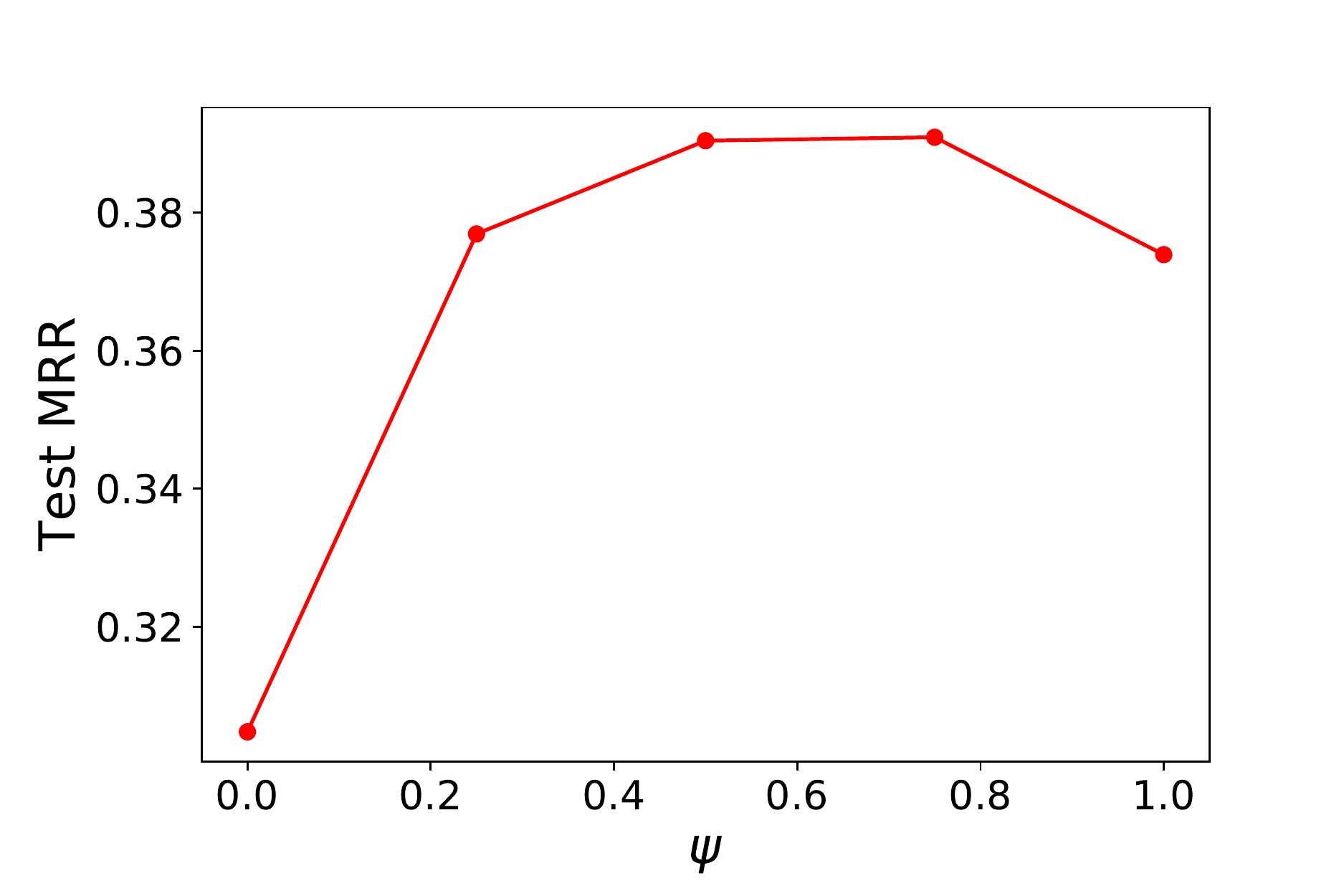}}
~~~~\hspace*{-0.5cm}
   \subfloat[]{%
   \includegraphics[width=0.3\textwidth]{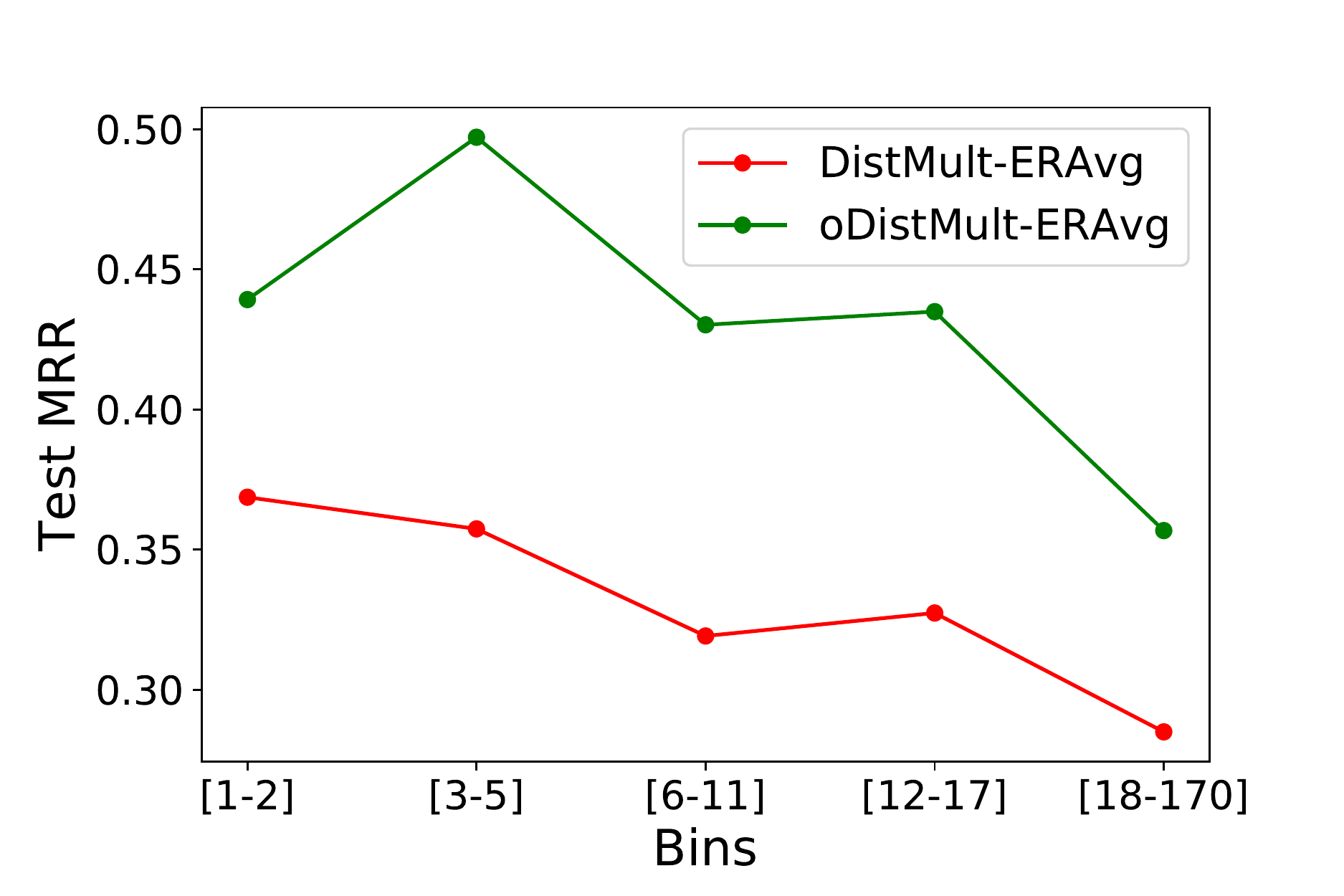}}
~~~~\hspace*{-0.5cm}
   \subfloat[]{%
   \includegraphics[width=0.3\textwidth]{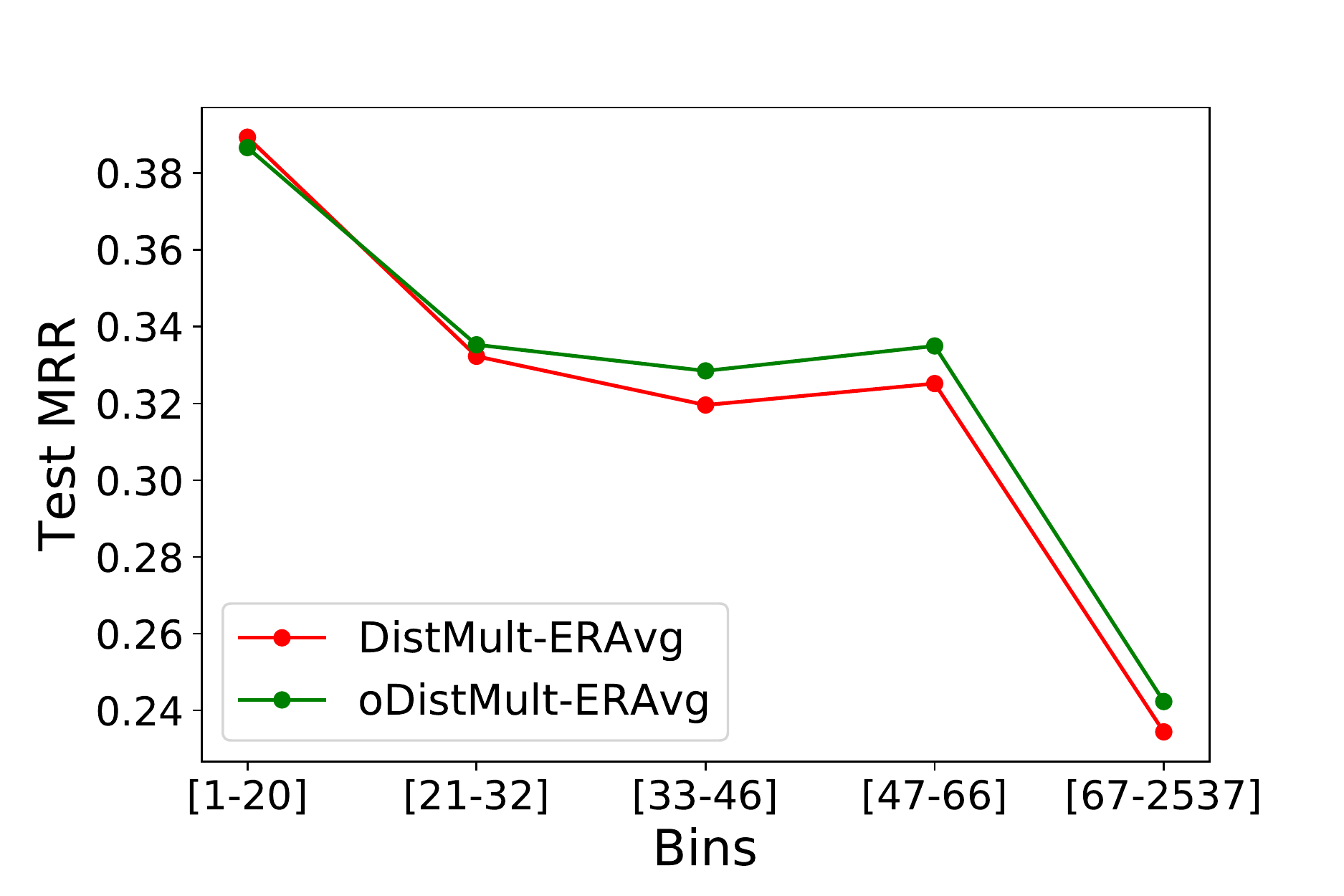}} %
    }
   \caption{%
   \label{fig:psi-bin-results} %
   (a) The test MRR of oDistMult-ERAvg on oWN18RR for different values of $\psi$ (introduced in Algorithm~\ref{algo:ind-training}). (b) and (c) Test MRR of DistMult-ERAvg and oDistMult-ERAvg on oWN18RR and oFB15k-237 for different bins (the bins are presented in Figure~\ref{fig:dataset-neighbours}).
   }
\end{figure*}

\subsection{Results}
According to the results on oWN18RR and oFB15k-237 reported in Table~\ref{results-table}, in almost all cases, using Algorithm~\ref{algo:ind-training} for training as opposed to Algorithm~\ref{algo:training} results in a boost of performance. Recall that the models whose names start with an ``o'' use Algorithm~\ref{algo:ind-training} and the models without ``o'' correspond to the variants where Algorithm~\ref{algo:training} is used instead. On oWN18RR, for instance, oDistMult-ERAvg and oDistMult-LS achieve $28\%$ and $16\%$ improvement in terms of filtered MRR compared to DistMult-ERAvg and DistMult-LS respectively. The margins of improvements on oFB15k-237 are smaller as oFB15k-237 is generally a more challenging dataset compared to oWN18RR and it is more difficult to make progress on.
We believe the reason for the observed boost when using Algorithm~\ref{algo:ind-training} is mainly because the train and test procedures become more consistent compared to when Algorithm~\ref{algo:training} is used. 

Furthermore, it can be observed that the proposed oDistMult-ERAvg and oDistMult-LS models outperform the other baselines. We believe the reason for the poor performance of RGCN-D on oWN18RR is because the out-of-sample entities have few neighbors (see Figure~\ref{fig:dataset-neighbours}(a)) and the degree information (used as initial features) is not discriminative enough\footnote{We tried a variant of RGCN without self-loops (similar to the model in \cite{hamaguchi2017knowledge}) but obtained similar results as RGCN-D.}. Between the two proposed models, the winner is dataset-dependant with oDistMult-LS performing slightly better on oWN18RR and oDistMult-ERAvg showing better performance on oFB15k-237. DistMult-LS also outperforms DistMult-LS-U shedding light on the importance of the normalization in Equation~\eqref{eq:ls}.

\textbf{Selecting $\mathbf{\psi}$:} For the results in Table~\ref{results-table}, we set the value of $\psi$ to $0.5$ (see Algorithm~\ref{algo:ind-training} for the usage of $\psi$). Here, we explore different values for $\psi$ to see how it affects the performance. Figure~\ref{fig:psi-bin-results}(a) shows the test MRR of oDistMult-ERAvg on oWN18RR for different values of $\psi$. When $\psi=0$ (corresponding to using the standard transductive training algorithm presented in Algorithm~\ref{algo:training}), the performance is poor. As soon as $\psi$ becomes greater than zero, we observe a substantial boost in performance. The performance keeps increasing as $\psi$ increases until reaching a plateau and then it goes down when $\psi=1$ corresponding to a training procedure where for each triple, one entity is always treated as out-of-sample. We repeated the experiment with other models and on other datasets and observed similar behavior. We believe one reason why we observe a better performance for $0<\psi<1$ compared to $\psi=1$ is that when $0<\psi<1$, the model is encouraged to learn embeddings that do well for both transductive and out-of-sample prediction tasks with the transductive task acting as an auxiliary task (and possibly as a regularizer) helping the embeddings capture more information.

\textbf{Neighbor-size effect:} Out-of-sample entities appear in a different number of triples. Figure~\ref{fig:dataset-neighbours} shows statistics for oWN18RR and oFB15k-237 on the number of triples used to learn the embedding for the out-of-sample entity in each query in the test set. To test how this number affects the models, we divided our test queries into 5 bins of (approximately) equal size as shown by the bar colors in Figure~\ref{fig:dataset-neighbours} and measured the test MRR on each bin. According to the results for oDistMult-ERAvg and DistMult-ERAvg, presented in Figure~\ref{fig:psi-bin-results}(b,c), oDistMult-ERAvg almost consistently outperforms DistMult-ERAvg on all (except one) bins. For both models, as the number of triples from which we learn the embedding for out-of-sample entities increases, the performance deteriorates, highlighting a shortcoming of our averaging strategy used for aggregation. Future work can look into other aggregation functions (e.g., attention-based averaging).

\textbf{In-sample performance:} To measure how training with Algorithm~\ref{algo:ind-training} affects model performance for in-sample (aka transductive) link prediction, we compared DistMult and oDistMult-ERAvg on the original splits of WN18AM, the cleaned version of WN18RR \cite{hajimoradlou2020stay}. For this experiment, we used Adam optimizer \cite{kingma2014adam} and added a dropout of $0.5$ after the Hadamard product of the embeddings (before taking the sum of the features) in DistMult. We tuned both learning rate and weight decay from the set $\{0.0001, 0.001, 0.01, 0.1\}$. The results in Table~\ref{results-in-sample} indicate that training with our proposed algorithm does not deteriorate the performance for in-sample link prediction.

\begin{table}[t]
\begin{center}
\resizebox{\columnwidth}{!}{%
\begin{tabular}{ccccc}
Model & MRR & Hit@1 & Hit@3 & Hit@10 \\ \hline
DistMult & 0.4498 & 0.4179 & 0.4614 & 0.5099 \\
oDistMult-ERAvg & 0.4483 & 0.4072 & 0.4711 & 0.5210
\end{tabular}
}
\caption{\label{results-in-sample} In-sample link prediction results on a cleaned version of WN18RR named WN18AM (for details, see \cite{hajimoradlou2020stay}). Although oDistMult-ERAvg has been trained for out-of-sample reasoning, its performance on in-sample reasoning is almost as good as DistMult.}
\end{center}
\end{table}

\section{Conclusion}
We studied out-of-sample representation learning for non-attributed multi-relational graphs - a problem that is surprisingly poorly studied. We created two benchmarks for this task and outlined the procedure we followed for creating these datasets to facilitate the creation of more datasets in the future. We also developed several baselines, a new training algorithm, and two aggregation models for out-of-sample representation learning. Future work includes developing new training strategies, testing other aggregation functions, combining the aggregation functions with other transductive models, extending out-of-sample reasoning to temporal KG completion and knowledge hypergraph completion (e.g., extending the proposed training algorithm and aggregation functions to the temporal or hypergraph versions of DistMult or SimplE \cite{goel2020diachronic,fatemi2019knowledge}) transferring the knowledge learned over one graph to a new graph with new entities (similar to \cite{zhang2020inductive,teru2019inductive}), studying the similarities and differences between out-of-sample representation learning and out-of-vocabulary word embedding, and testing the proposed models on relational domains other than knowledge graphs.

\bibliographystyle{acl_natbib}
\bibliography{MyBib}

\end{document}